# Entropy and Syntropy in the Context of Five-Valued Logics

Vasile Pătraşcu

Department of Informatics Technology, Tarom Company
Bucharest, Romania
email: patrascu.v@gmail.com

**Abstract**. This paper presents a five-valued representation of bifuzzy sets. This representation is related to a five-valued logic that uses the following values: true, false, inconsistent, incomplete and ambiguous. In the framework of five-valued representation, formulae for similarity, entropy and syntropy of bifuzzy sets are constructed.

## 1   Introduction

Here we propose a measure of bifuzzy entropy which is a generalization of fuzzy entropy. Complementary, a bifuzzy syntropy is also proposed. The proposed bifuzzy entropy is the similarity measure between the considered elements and its complement while the bifuzzy syntropy is the dissimilarity measure [15]. The measures of similarity/dissimilarity are computed in the framework of five-valued knowledge representation. The presented formulae describe entropy/syntropy for a single element belonging to a bifuzzy set. For a set we must sum the entropy/syntropy of all its elements. The proposed measure of entropy reaches its maximum for elements for which membership function is equal to non-membership function. The structure of entropy offers the advantage we immediately are able to indicate possible reasons of the missing information: the inconsistency, the incompleteness, the lack of distinguish between membership/non-membership and their negations. The paper has the following structure: section 2 presents the bifuzzy sets and their particular forms. Section 3 enumerates some existent four and five logics having partial order structure. Section 4 presents a five-valued logic that uses three neutral values: inconsistent, incomplete and ambiguous. Section 5 presents a five-valued representation of bifuzzy sets. Section 6 presents measures for similarity, entropy and syntropy of bifuzzy sets. Finally, the conclusions are presented in section 7.

## 2 Bifuzzy Set and Its Particular Forms

Let *X* be a crisp set. In the framework of Zadeh theory [17], a *fuzzy set A* is defined by the membership function $\mu_A : X \to [0,1]$. The non-membership function $\nu_A : X \to [0,1]$ is obtained by negation and thus both functions define a partition of unity, namely:

$$\mu_A + \nu_A = 1 \tag{2.1}$$

Atanassov has extended the fuzzy sets to the *intuitionistic fuzzy sets* [1]. Atanassov has relaxed the condition (2.1) to the following inequality:

$$\mu_A + \nu_A \leq 1 \tag{2.2}$$

He has used the third function, the intuitionistic fuzzy index $\pi_A$ that verifies the equality:

$$\pi_A = 1 - \mu_A - \nu_A \tag{2.3}$$

Similarly, we can consider instead of (2.1) the following condition:

$$\mu_A + \nu_A \geq 1 \tag{2.4}$$

Thus, we obtain the *paraconsistent fuzzy set* [14]. One can define the index of contradiction:

$$\kappa_A = \mu_A + \nu_A - 1 \tag{2.5}$$

More generally, in this paper, we will consider as *bifuzzy set* a set $A$, defined by two functions totally independent $\mu_A : X \to [0,1]$ and $\nu_A : X \to [0,1]$. We consider the following two parameters:

*the net truth*: $\quad\quad\quad\quad \tau_A = \mu_A - \nu_A \tag{2.6}$

*the definedness* $\quad\quad\quad\quad \delta_A = \mu_A + \nu_A - 1 \tag{2.7}$

When $\delta_A > 0$ the information is inconsistent (overdefined) and when $\delta_A < 0$ the information is incomplete (undefined). In addition, from (2.6) and (2.7) it results the following inequality:

$$|\tau_A| + |\delta_A| \leq 1 \tag{2.8}$$

We must observe that $|\tau|$ is a distance between $\mu$ and $\nu$. Also, $|\delta|$ is a distance between $\mu$ and $1 - \nu$. One can generalize the formulae (2.6) and (2.7). Let there be $a, b \in [0,1]^2$ and $\bar{a} = 1 - a$, $\bar{b} = 1 - b$. We can consider the following equality:

$$\frac{a \cdot \bar{b}}{1 - a \cdot b} + \frac{\bar{a} \cdot b}{1 - a \cdot b} + \frac{\bar{a} \cdot \bar{b}}{1 - a \cdot b} = 1 \tag{2.9}$$

If $a = |\mu - \nu|$ and $b = |\mu + \nu - 1|$ one obtains a new distance:

$$D(\mu,\nu) = \frac{a \cdot \bar{b}}{1 - a \cdot b} = \frac{|\mu - \nu| \cdot (1 - |\mu + \nu - 1|)}{1 - |\mu - \nu| \cdot |\mu + \nu - 1|} \tag{2.10}$$

and its equivalent form:

$$D(\mu, 1-\nu) = \frac{\bar{a} \cdot b}{1 - a \cdot b} = \frac{|\mu + \nu - 1| \cdot (1 - |\mu - \nu|)}{1 - |\mu - \nu| \cdot |\mu + \nu - 1|}$$

It results, new formulae for $\tau$ and $\delta$:

$$\tau = \frac{1-|\mu+\nu-1|}{1-|\mu-\nu|\cdot|\mu+\nu-1|}\cdot(\mu-\nu) \tag{2.11}$$

$$\delta = \frac{1-|\mu-\nu|}{1-|\mu-\nu|\cdot|\mu+\nu-1|}\cdot(\mu+\nu-1) \tag{2.12}$$

## 3   Four and Five-valued Logics Partial Ordered

Four and five-valued logics are a good deal rarer in the literature than three-valued ones. Goddard and Routley considered a logic with following values: true $T$, false $F$, nonsignificant $N$ and incomplete $I$ [9]. Belnap has defined his well known logic using the values true $T$, false $F$, none $N$ and both $B$ [3]. Bergstra, Bethke, Rodenburg proposed a four-valued logic that uses the values: true $T$, false $F$, divergent $D$ and meaningless $M$ [4]. Delgado, Sanchez and Vila considered a four-valued logic that uses the values: true $T$, false $F$, unknown $U$ and contradiction $C$ [6]. Their logic is different from Belnap one. Ferrreira has defined a five-valued logic. The five values correspond to the following semantics: true $tt$, false $ff$, inconsistent $ii$, undefined or unknown $uu$ and (possibly) known but consistent $kk$ [7]. He used two-letter symbols because, he considered that symbols such as $t,f,u,i$ are commonly used in mathematics and other sciences. Lewis introduces a five-valued logic with the values: true $1$, false $0$, up $0/1$, down $1/0$ and ambiguous $\frac{1}{2}$ [12].

Levis developed the five valued logic from Metze's four valued logic [20]. Bergstra and Ponse have defined a five-valued logic de uses the values: true $T$, false $F$, meaningless $M$, choice $C$ and divergent $D$ [5]. In this paper is presented a five-valued logic that differs from all these enumerated above.

## 4  Five-valued Logic Based on Inconsistency, Incompleteness and Ambiguity

In the framework of this logic we will consider the following five values: *true* $t$, *false* $f$, *incomplete* (*undefined*) $u$, *inconsistent* (*contradictory, overdefined*) $c$, and *ambiguous* (*equidistant*) $i$. Tables 1, 2, 3, 4, 5, 6 and 7 show the basic operators in this logic.

Entropy and Syntropy in the Context of Five-Valued Logics

**Table 1**: The disjunction

| ∪ | t | i | u | c | f |
|---|---|---|---|---|---|
| t | t | t | t | t | t |
| i | t | i | i | i | i |
| u | t | i | u | i | u |
| c | t | i | i | c | c |
| f | t | i | u | c | f |

**Table 2**: The conjunction

| ∩ | t | i | u | c | f |
|---|---|---|---|---|---|
| t | t | i | u | c | f |
| i | i | i | i | i | f |
| u | u | i | u | i | f |
| c | c | i | i | c | f |
| f | f | f | f | f | f |

**Table 3**: The complement

|   | ~ |
|---|---|
| t | f |
| i | i |
| u | u |
| c | c |
| f | t |

**Table 4**: The negation

|   | ¬ |
|---|---|
| t | f |
| i | i |
| u | c |
| c | u |
| f | t |

**Table 5**: The dual

|   | ≈ |
|---|---|
| t | t |
| i | i |
| u | c |
| c | u |
| f | f |

Vasile Pătraşcu

**Table 6**: The implication

| → | t | i | u | c | f |
|---|---|---|---|---|---|
| t | t | i | u | c | f |
| i | t | i | i | i | i |
| u | t | i | u | i | u |
| c | t | i | i | c | c |
| f | t | t | t | t | t |

**Table 7**: The equivalence

| ↔ | t | i | u | c | f |
|---|---|---|---|---|---|
| t | t | i | u | c | f |
| i | i | i | i | i | i |
| u | u | i | u | i | u |
| c | c | i | i | c | c |
| f | f | i | u | c | t |

## 5. Transformation from Bivalued Knowledge Representation to a Five-Valued One

In order to obtain a multi-valued representation of bifuzzy values, we can define a fuzzy partition on the unit square. Taking into account the geometrical form of the square and its symmetry the first values of number of classes could be 4, 5 or 9. For 9 classes, the model is too complicated. For 4 classes we can not consider the center of the square, where the entropy of fuzzy sets has its maximum value. Because of this reasons, in this paper it will be defined a five-valued fuzzy partition. This partition is related to the five-valued logic presented in this paper. Let there be a bifuzzy value $(\mu,\nu) \in [0,1]^2$. From (2.6) and (2.7) one obtains $\tau$ and $\delta$. We will define the following 5 parameters:

*index of truth*: $\qquad t = \tau_+$ $\qquad\qquad$ (5.1)

*index of falsity*: $\qquad f = \tau_-$ $\qquad\qquad$ (5.2)

*index of inconsistency*: $\qquad c = \delta_+$ $\qquad\qquad$ (5.3)

*index of incompleteness*: $\qquad u = \delta_-$ $\qquad\qquad$ (5.4)

*index of ambiguity*: $\qquad i = 1 - |\tau| - |\delta|$

where $x_+ = \max(x,0)$ and $x_- = \max(-x,0)$.

It results the following five-valued fuzzy partition of unity:
$$t + f + u + c + i = 1 \qquad (5.5)$$

We must observe that $t$ is related to the point $(1,0)$, $f$ is related to the point $(0,1)$, $u$ is related to the point $(0,0)$, $c$ is related to the point $(1,1)$ and $i$ is related to the point $(0.5, 0.5)$. Let there be $x = (t, c, u, f) \in [0,1]^4$, For this kind of vectors, one defines the union, the intersection, the complement, the negation, the complement and the dual operators.

*The Union*: For two vectors $x_1, x_2 \in [0,1]^4$ one defines the union $x_1 \cup x_2$ by the formula:

$$\begin{aligned} x_1 \cup x_2 = (&t_1 \vee t_2, \\ &(c_1 + f_1) \wedge (c_2 + f_2) - f_1 \wedge f_2, \\ &(u_1 + f_1) \wedge (u_2 + f_2) - f_1 \wedge f_2, \\ &f_1 \wedge f_2) \end{aligned} \quad (5.6)$$

*The Intersection*: For two vectors $x_1, x_2 \in [0,1]^4$ one defines the intersection $x_1 \cap x_2$ by the formula:

$$\begin{aligned} x_1 \cap x_2 = (&t_1 \wedge t_2, \\ &(c_1 + t_1) \wedge (c_2 + t_2) - t_1 \wedge t_2, \\ &(u_1 + t_1) \wedge (u_2 + t_2) - t_1 \wedge t_2, \\ &f_1 \vee f_2) \end{aligned} \quad (5.7)$$

In formulae (5.6) and (5.7), the symbols "$\vee$" and "$\wedge$" represent any couple of Frank t-conorm, Frank t-norm.

If we want to preserve de properties $u \cdot c = 0$ it is necessary to choose only and only $\vee = \max$ and $\wedge = \min$ in formulae (5.6) (5.7).

*The Complement*: For $x = (t, c, u, f) \in [0,1]^4$ one defines the complement $x^c$ by formula:

$$x^c = (f, c, u, t) \quad (5.8)$$

*The Negation*: For $x = (t, c, u, f) \in [0,1]^4$ one defines the negation $x^n$ by formula:

$$x^n = (f, u, c, t) \quad (5.9)$$

*The Dual*: For $x = (t, c, u, f) \in [0,1]^4$ one defines the dual $x^d$ by formula:

$$x^d = (t, u, c, f) \quad (5.10)$$

In the set $\{0,1\}^4$ there are five vectors having the form $x = (t, c, u, f)$, which verify the condition $t + f + c + u \leq 1$: $T = (1,0,0,0)$ (*True*), $F = (0,1,0,0)$ (*False*), $C = (0,0,1,0)$ (*Inconsistent*), $U = (0,0,0,1)$ (*Incomplete*) and $I = (0,0,0,0)$ (*Ambiguous*). Using the operators defined by (5.6), (5.7), (5.8), (5.9) and (5.10), the same truth table results as seen in Tables 1, 2, 3, 4, 5.

## 6   Similarity, Entropy and Syntropy of Bifuzzy Sets

For a non-crisp set, uncertainty results from the imprecise boundaries or it results from the lack of crisp distinction between the elements belonging and not belonging

Entropy and Syntropy in the Context of Five-Valued Logics

to a set. The uncertainty of fuzzy sets is called fuzziness. A measure of fuzziness is the entropy first mentioned by Zadeh [21]. Kaufman [10] proposed to measure a degree of fuzziness of any fuzzy set $A$ by a metric distance between its membership function and the membership function of its nearest crisp set. Another way given by Yager [16] was to view a degree of fuzziness in term of a lack of distinction between the fuzzy set and its complement. De Luca and Termini [13] introduced some requirements for entropy of fuzzy sets. Szmidt and Kacprzyk have extended these requirements from fuzzy sets to intuitionistic fuzzy sets. Kosko [11] proposed to measure the fuzzy entropy by the ratio between the distance to the nearest crisp element and the distance to the farthest crisp element. But this formula is nothing else then a similarity measure between the considered elements and its complement. This formula will be used in order to define the bifuzzy entropy. Hence in fact the measure is not unique, because of the particular form of similarity that is used. Moreover, such entropy is dependent on the representation of element $x$. Firstly, we define de similarity of bifuzzy values. Let there be a bifuzzy value $x = (\mu, \nu) \in [0,1]^2$ and its five-valued representation $(t, f, c, u, i)$. Because $t + f + c + u + i = 1$ one can use the Bhattacharyya similarity for two bifuzzy values [2].

$\forall x_1, x_2 \in [0,1]^2$

$$S(x_1, x_2) = \sqrt{t_1 t_2} + \sqrt{f_1 f_2} + \sqrt{c_1 c_2} + \sqrt{u_1 u_2} + \sqrt{i_1 i_2} \qquad (6.1)$$

Having the similarity, one can define the entropy. Therefore we get:

$$x = (\mu, \nu) = (t, f, c, u, i) \qquad (6.2)$$

$$x^c = (\nu, \mu) = (f, t, c, u, i) \qquad (6.3)$$

$$S(x, x^c) = \sqrt{t \cdot f} + \sqrt{f \cdot t} + \sqrt{c \cdot c} + \sqrt{u \cdot u} + \sqrt{i \cdot i} \qquad (6.4)$$

$$e(x) = S(x, x^c) = c + u + i \qquad (6.5)$$

This is the entropy of bifuzzy value $x = (\mu, \nu)$ and represents the total uncertainty. The defined entropy has three components related to inconsistency, incompleteness and ambiguity. In order to have a complete description of entropy we define the following vector:

$$\vec{e} = (c, u, i) \qquad (6.6)$$

Our vector approach of bifuzzy entropy is close to that defined by Grzegorzewski and Mrowka for intuitionistic fuzzy sets [8]. Also, we define the syntropy [18],[19]:

$$\gamma = t + f \qquad (6.7)$$

We must underline that $\gamma$ is a measure for syntropy because it is the negation of entropy. Also this parameter is known as negentropy. The syntropy has two components: the index of truth and index of falsity. In addition, we define the following vector of syntropy:

$$\vec{\gamma} = (t, f) \qquad (6.8)$$

Entropy (6.5) and syntropy (6.7) verify the equalities:

$\forall x, y \in [0,1]^2 \qquad e(x \cup y) + e(x \cap y) = e(x) + e(y)$

$\qquad \qquad \gamma(x \cup y) + \gamma(x \cap y) = \gamma(x) + \gamma(y)$

where $\cup$ and $\cap$ are defined by (5.6) and (5.7).
Using (2.11), (2.12) one obtains the following particular forms for $e$ and $\gamma$:

$$e = \frac{1-|\mu-\nu|}{1-|\mu-\nu|\cdot|\mu+\nu-1|}$$

$$\gamma = \frac{|\mu-\nu|\cdot(1-|\mu+\nu-1|)}{1-|\mu-\nu|\cdot|\mu+\nu-1|}$$

We consider the points $x=(0.5,0.5)$ and $y=(0,0)$. One obtains: $e(x)=e(y)=1$, $\vec{e}(x)=(0,0,1)$, $\vec{e}(y)=(0,1,0)$ where $\vec{e}=(c,u,i)$. Analyzing the scalars $e(x)$ and $e(y)$, it results that both points have the same entropy but the nature of entropy is not the same. Analyzing the vector entropies $\vec{e}(x)$ and $\vec{e}(y)$, it results that the first point has as source of its entropy, the ambiguity while the second has the incompleteness.

## 7 Conclusions

In this paper, a method was presented regarding multi-valued knowledge representation. Also, a five-valued logic was presented based on five values: true, false, incomplete, inconsistent and ambiguous. Using this logic, new representations are obtained for bifuzzy set. These new representations supply formulae for entropy and syntropy of bifuzzy sets. This approach offers a vector structure with three components for entropy and with two components for syntropy. However, the proposed measure is consistent with the classical entropy for standard fuzzy sets.

Vasile Pătraşcu